\documentclass[letterpaper]{article} %
\usepackage{iclr2026_conference,times}

\usepackage{amsmath,amsfonts,bm}

\def\eqref#1{equation~\ref{#1}}

\def\1{\bm{1}}

\DeclareMathAlphabet{\mathsfit}{\encodingdefault}{\sfdefault}{m}{sl}
\SetMathAlphabet{\mathsfit}{bold}{\encodingdefault}{\sfdefault}{bx}{n}

\usepackage{hyperref}
\usepackage{url}

\usepackage{times}
\usepackage{latexsym}
\usepackage{booktabs} 
\usepackage{multirow}
\usepackage{adjustbox}
\usepackage{subcaption}
\usepackage{wrapfig}

\title{Decoding Partial Differential Equations: \\ Cross-Modal Adaptation of Decoder-only Models to PDEs}

\author{Paloma Garc\'ia-de-Herreros\textsuperscript{1} \,
  Philipp Slusallek\textsuperscript{1, 2} \, Dietrich Klakow\textsuperscript{1} \,
  Vagrant Gautam\textsuperscript{3}
 \vspace{3px} \\ \vspace{3px}
 \textsuperscript{1}Saarland University~~~ 
    \textsuperscript{2}DFKI~~~\textsuperscript{3}Heidelberg Institute for Theoretical Studies~~~\\
\small{\tt  pgherreros@lsv.uni-saarland.de \quad vagrant.gautam@h-its.org}
}

\iclrfinalcopy %
\begin{document}

\maketitle

\begin{abstract}
While large language models are primarily used on natural language tasks, they have also shown great promise when adapted to new modalities, e.g., for scientific machine learning tasks.
Most proposed approaches for such cross-modal adaptation of language models focus on encoder-only transformer model architectures, despite decoder-only architectures being far more popular for language tasks in recent years, and being trained at much larger scales.
This raises the question of how model architecture affects cross-modal adaptation approaches, and whether we can leverage the success of decoder-only models.
In this paper, we systematically compare encoder-only and decoder-only language models on cross-modal adaptation for time-dependent simulation tasks based on partial differential equations (PDEs).
We find that decoder-only models are far worse than encoder-only models, when existing approaches are applied unmodified.
In contrast to several other domains, scaling decoder-only models also does not help.
To enhance the performance of decoder-only models in this context, we introduce two novel approaches that mimic bidirectionality, Parallel Flipping and Sequence Doubling.
Both our methods improve overall performance using decoder-only models for all tasks and all cross-modal adaptation methods, closing the gap to encoder-only model performance.
We hope that our findings broaden the spectrum of models used on cross-modal adaptation tasks to further scientific machine learning.
\end{abstract}

\section{Introduction}

Pre-trained large language models (LLMs) have seen unprecedented improvements in processing natural language in recent years.
These models can then be adapted to new tasks using different approaches like fine-tuning or in-context learning. 
Recent work has even applied such methods to adapt models to data modalities unseen during pre-training, an approach termed cross-modal adaptation.
Cross-modal adaptation methods purport to take advantage of the knowledge, skills, or simply the already-optimized neural network of a pre-trained model, to adapt it to a new modality.
Cross-modal adaptation achieves competitive performance across a wide range of tasks, including detecting atrial cardiac disease from ECG recordings, and time-dependent simulation tasks based on partial differential equations \citep{lu2022frozen,shen2023cross,pmlr-v235-ma24d,shen2024ups}.
These approaches can be of great utility for scientific machine learning tasks, currently used for tasks such as seismic monitoring \citep{wang2025seismollm} and time series forecasting \citep{liu2025calf}.

However, it is unclear how general cross-modal adaptation methods are, since few studies vary the originally proposed configurations.
For example, most approaches are based on encoder-only models, even though decoder-only models are by far the more popular transformer-based model architecture \citep{Vaswani+2017} for natural language tasks; they scale impressively, mimic human-like language convincingly, and can be used for a wide variety of tasks.
Through better, more general representations of natural language, today's best decoder-only models may well provide a better starting point for cross-modal adaptation.

In our work, we attempt to leverage the potential of decoder-only models to broaden the range of models available for cross-modal adaptation.
First, we directly apply two existing cross-modal adaptation methods to encoder-only and decoder-only models, and find that decoder-only models perform much worse (Section \ref{sec:architecture}).
Next, we test whether scaling up decoder-only models helps performance, but find that it does not (Section \ref{sec:scaling}).

\begin{figure*}
    \centering
    \includegraphics[width=\linewidth]{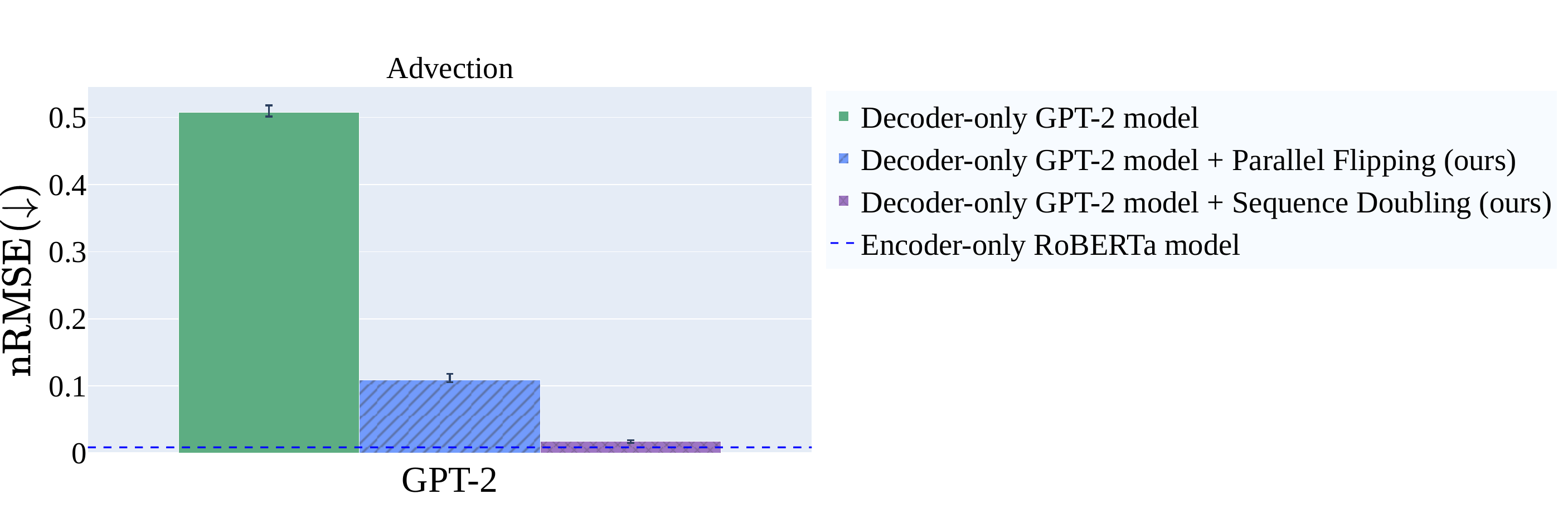}
    \caption{Cross-modal adaptation of \textsc{GPT-2}, a decoder-only model, with ORCA-based adaptation on the Advection dataset of time-dependent PDE simulation compared to the original performance of \textsc{RoBERTa} with the same cross-modal adaptation approach as a baseline. Although the original setup shows high error, our proposed methods (Parallel Flipping and Sequence Doubling) close the gap to encoder-only model performance.}
    \label{fig:one}
\end{figure*}

We hypothesize that the reasons for the lack of success of these approaches are due to autoregressive attention over the input, as well as how the outputs are computed, which is by averaging the representations of the last hidden layer, rather than generating outputs, as in natural language.
Addressing these issues, we introduce two different methods to improve cross-modal adaptation with decoder-only models by simulating bidirectionality (Section \ref{sec:methods}):

\paragraph{Parallel Flipping: } In parallel, we run the original setup and the same pipeline with the data inverted, and we then combine the predictions by taking the second half of both of them.

\paragraph{Sequence Doubling: } We concatenate every sequence in the original data with itself before introducing it to the model, and use only the second half of the last hidden layer to compute predictions.

Both methods extend the sequence context that models can access at different points, and results show that our methods outperform the original setup for all tasks and cross-modal adaptation methods, closing the gap to encoder-only models (Section \ref{sec:results}), as shown in Figure \ref{fig:one}.
Each method comes with particular tradeoffs, making this a promising direction for future work, which we discuss in Section \ref{sec:discussion}.
We hope that our findings broaden the spectrum of models used for cross-modal adaptation and further the field of scientific machine learning.

\section{Related Work}

\subsection{Machine Learning and Partial Differential Equations}

Solving partial differential equations has been considered a classical machine learning problem, given its scientific significance, with task-specific models proposed including the physics-informed neural network \citep[PINN;][]{pinn}, Fourier neural operator \citep[FNO;][]{fno}. In recent years, foundational PDE models like PDE-FM \citep{soares2026towards}, POSEIDON \citep{herde2024poseidon}, Walrus \citep{mccabe2025walruscrossdomainfoundationmodel}, and UNISOLVER \citep{zhou2025unisolver} have been introduced.
All of these approaches focus on custom-built machine learning solutions to solve PDEs.

In contrast, other approaches try to leverage other types of machine learning models (such as large language models) to enhance analytical PDE approximations \citep{bhatnagar2025equations}, find the analytical solution to differential equations \citep{surkov2024application,zakharov2025agdes,xiao2026llm4fluid}.
Some work has even used LLMs to generate code \citep{li2025codepde} to simulate the data modeled by certain PDEs, as proposed in \citet{takamoto2022pdebench}, while other work uses in-context learning to simply prompt LLMs for PDE simulation tasks \citep{bao2026texttrained}
Our work differs from the above by using cross-modal adaptation of LLMs to solve these tasks, similar to \citet{shen2023cross} and \citet{shen2024ups}.
This data consists of time series predictions of continuous observations over a space domain, similar to other scientific machine learning data such as Satellite \citep{petitjean2012satellite} and MegaFlow2D \citep{10.1145/3576914.3587552}, which our PDE-focused work could also be relevant to.

\subsection{Cross-Modal Adaptation}

Although language model adaptation is an important feature within natural language applications (across domains, languages, varieties, for particular use cases, etc.), a more extreme approach known as cross-modal adaptation has recently been introduced, which involves adapting models to new modalities unseen by the model during pre-training.
The concept of modality is not well-defined in this literature, but the key idea is to adapt models to a data modality that is entirely different from the modality that a model was pre-trained on, and which should, in theory, have no connection, unlike speech and text representing the same utterance.
Most of the proposed methods focus on the fine-tuning stage of large language models.
These methods include Frozen Pretrained Transformers \citep[FPT;][]{lu2022frozen}, ORCA \citep{shen2023cross}, Patch Replacement \citep[PaRe;][]{pmlr-v235-cai24c}, Modality kNowledge
Alignment \citep[MoNA;][]{pmlr-v235-ma24d}, Unified PDE Solver  \citep[UPS;][]{shen2024ups}, and more.
All these methods take advantage of model pre-training, to minimize the amount of fine-tuning necessary to adapt it to a new modality.
These techniques have a lot of potential to be used for various scientific machine learning tasks, and some practical applications have also been presented, including seismic monitoring \citep{wang2025seismollm}, time series forecasting \citep{liu2025calf}, and physics simulations \citep{ohana2024well}.

\subsection{Architecture Differences}

Virtually all modern large language models are based on the transformer architecture, consisting of an encoder and a decoder \citep{Vaswani+2017}.
This encoder-decoder architecture has been replaced by encoder-only architectures such as \textsc{BERT} \citep{DBLP:journals/corr/abs-1810-04805}, as well as decoder-only architectures such as \textsc{GPT} \citep{radford2019language}, the latter two of which are more popular in modern natural language processing applications.
Due to differences between the architectures, including pre-training objectives and attention mechanisms, several works compare them, finding differences in phenomena such as pronoun use \citep{10.1162/tacl_a_00719} and various linguistic probes \citep{waldis-et-al-2024-holmes}.
In cross-modal adaptation, however, there have been no systematic architectural comparisons, to the best of our knowledge.
Some papers, like us, also try to close the gap between encoder-only and decoder-only models in various contexts, including language model embeddings \citep{behnamghader2024llmvec,springer2025repetition} and cognitively plausible language modelling \citep{charpentier-samuel-2024-bert}.

\section{Experimental Setup}\label{sec:setup}

To evaluate the effects of model architecture and scaling on cross-modal adaptation with partial differential equation data, we experiment with several models, scales, and cross-modal adaptation methods as described below.

\subsection{Methods}

We choose two popular methods for cross-modal adaptation in the literature -- Frozen Pretrained Transformers (FPT) \citep{lu2022frozen} and ORCA \citep{shen2023cross}.
In both cases, a task-specific embedder and predictor are created to account for mismatches in dimensions between the target modality data and the original model.
Then, FPT adapts the pre-trained models to new tasks by fine-tuning only the input and output layers, as well as the layer normalization parameters.
ORCA, on the other hand, first trains the embedder on its own, minimizing the Optimal Transport Dataset Distance (OTDD) \citep{alvarez2020geometric} between the target task dataset and a pre-selected proxy dataset.
After this, all parameters are trained on the target task dataset.
We use ORCA's implementation for both ORCA and FPT, with the same hyperparameters.

\subsection{Models} 

We select \textsc{RoBERTa-BASE} \citep{DBLP:journals/corr/abs-1907-11692} and \textsc{BERT} as our encoder-only models, following ORCA~\citep{shen2023cross}, and \textsc{GPT-2} \citep{radford2019language} and \textsc{Pythia} \citep{biderman2023pythia} as our decoder-only models, since \textsc{GPT-2} is used in \citet{lu2022frozen} and \textsc{Pythia-160M} has a large range of model sizes. All of these models have similar sizes (respectively, 125M, 110M, 137M, and 160M parameters). For the scaling experiments, we consider the larger versions of the \textsc{GPT-2} family: \textsc{GPT-2 Medium}  (380M), \textsc{GPT-2 Large}  (812M), and \textsc{GPT-2 XL} (1.61B), as well as the \textsc{Pythia} family: \textsc{Pythia-14M}, \textsc{Pythia-70M}, \textsc{Pythia-410M}, \textsc{Pythia-1B}, and \textsc{Pythia-1.4B}. We do not consider larger \textsc{Pythia} model sizes to keep the comparison with the \textsc{GPT-2} family fair.

\subsection{Datasets}

We use four different datasets of time-dependent simulation tasks based on partial differential equations:
Advection, Diffusion-Reaction, Diffusion-Sorption, and Navier-Stokes, all taken from PDEBench \citep{takamoto2022pdebench}.
We follow the configurations in \citet{shen2023cross} as detailed in Appendix \ref{sec:appendix-pde}.
As in their work, we train the models to predict the last time step from the first time step, instead of a shorter one-step prediction or a multi-step rollout.

\paragraph{Proxy Datasets} In addition to the target dataset, the ORCA method also requires a proxy dataset for training the embedder. For \textsc{RoBERTa-BASE}, we use
the original proxy dataset generated by \citet{shen2023cross} using CoNLL-2003.
We follow their approach with CoNLL-2000 to generate proxy datasets for the rest of the models, as we explain in detail in Appendix \ref{sec:appendix-proxy}.

\subsection{Evaluation Metric}

As in previous literature \citep{shen2023cross,pmlr-v235-ma24d,shen2024ups,pmlr-v235-cai24c,li2025codepde}, we report normalized Root Mean Squared Errors (nRMSE) for all tasks, as it is scale-independent.
As the metric is error-based, lower values are better, which we also note in all figure captions.
We report averages over five runs; given the high variance for some configurations, we show best (minimum) and worst (maximum) performance with error bars.

\section{Decoder-Only Models Perform Much Worse than Encoder-Only Models}\label{sec:architecture}

In this section, we experiment with two transformer architectures, encoder-only and decoder-only models, represented by \textsc{RoBERTa-BASE} and \textsc{BERT-BASE}, and \textsc{GPT-2} and \textsc{Pythia-160M}, respectively, plugged directly into the existing cross-modal adaptation approaches, FPT and ORCA.
Prior work generally assumes that pre-training results in better cross-modal adaptation performance, but we ablate for this factor as well by including randomly-initialized versions of these models.
This allows us to disentangle the effects of both architecture and pre-training.

\begin{figure*}[ht]
\begin{center}
\includegraphics[width=1.0\textwidth]{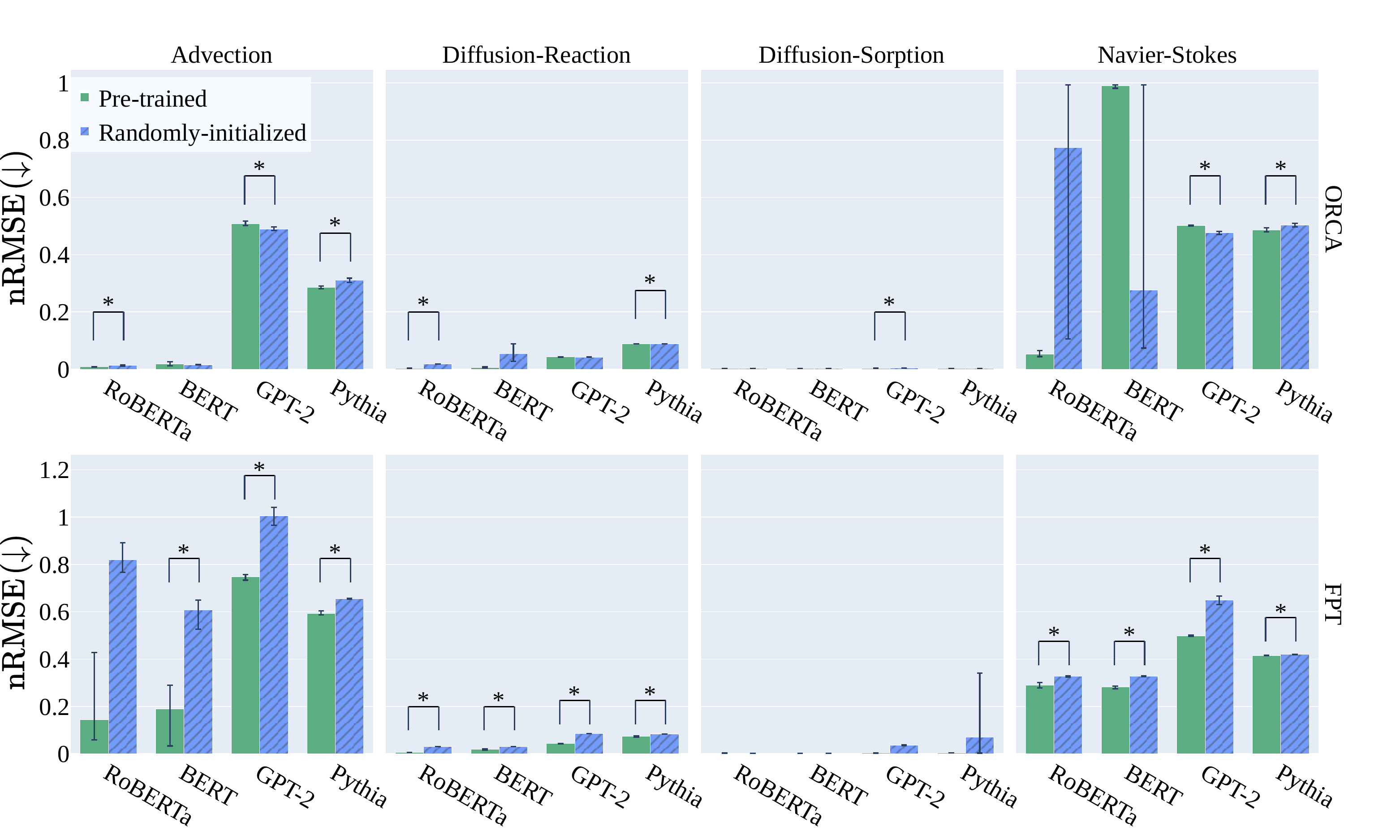}
\end{center}
\caption{Comparison of model performance with ORCA- (above) and FPT-based (below) cross-modal adaptation, using both pre-trained and randomly-initialized versions of encoder-only models (\textsc{RoBERTa}, \textsc{BERT}) and decoder-only models (\textsc{GPT-2}, \textsc{Pythia}). Performance is measured using nRSME, where lower is better; the plots show average performance over five random seeds, and the error bars represent the best and worst runs. Statistically significant differences between pre-trained and randomly-initialized models are denoted with $\ast$); a detailed explanation can be found in Appendix \ref{sec:appendix-statistic}. Since the scale used can make Diffusion-Reaction and Diffusion-Sorption plots hard to read, complementary plots can be found in Appendix \ref{sec:appendix-zoomin}.}
\label{fig:architectures}
\end{figure*}

\begin{figure*}[ht]
\begin{center}
\includegraphics[width=1.0\textwidth]{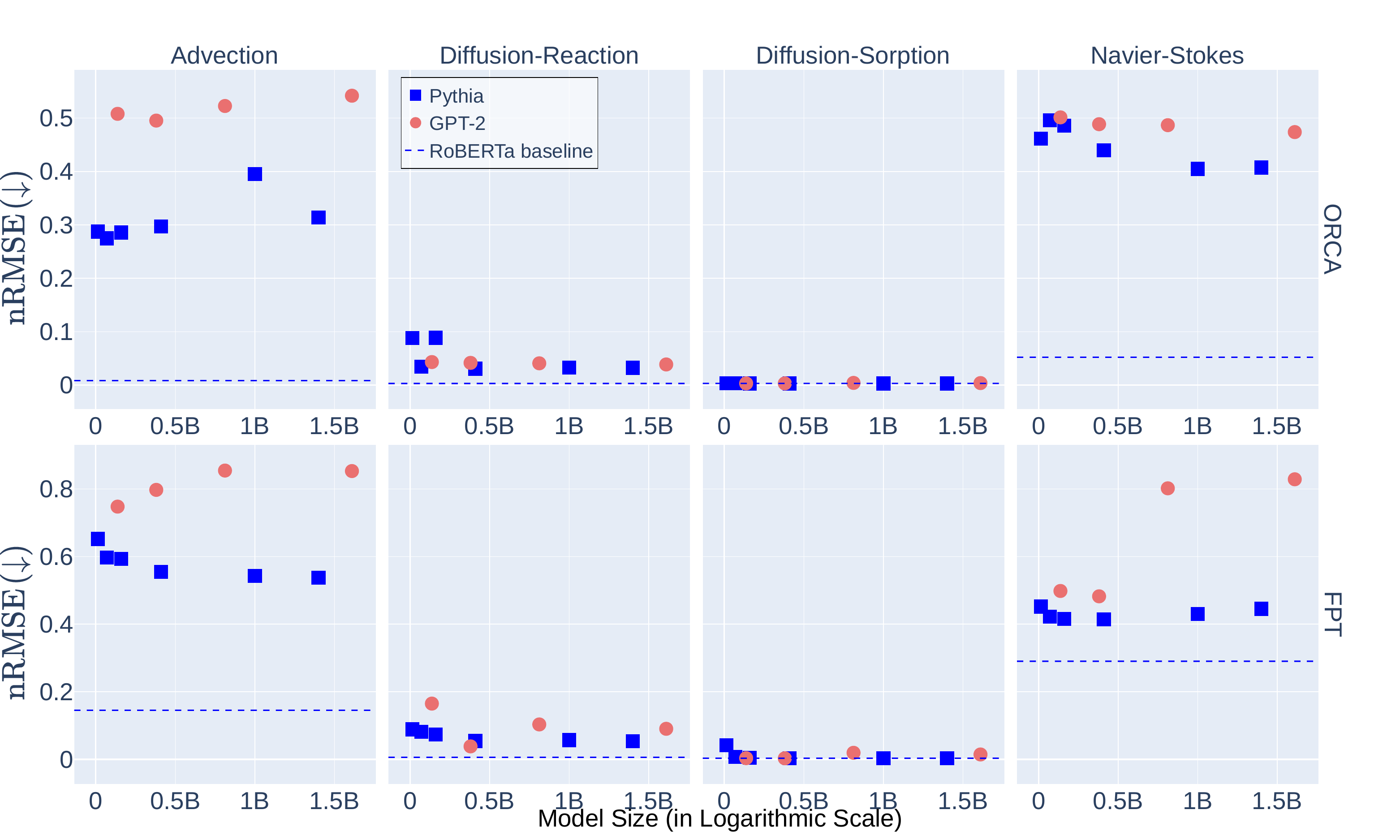}
\end{center}
\caption{Performance of different sizes of models of the \textsc{GPT-2} family and \textsc{Pythia} family using both ORCA \citep{shen2023cross} and FPT \citep{lu2022frozen}. The plots depict the average performance over five random seeds. Once again, performance is measured using nRSME, where lower is better. If scaling the models was improving the performance, downward trends could've been seen for the different model families. Since the scale used can make Diffusion-Reaction and Diffusion-Sorption plots hard to read, complementary plots can be found in Appendix \ref{sec:appendix-zoomin}.}
\label{fig:scaling}
\end{figure*}

We start by considering the performance of randomly-initialized versions of the models, to evaluate whether pre-training on language data actually helps at all with these tasks. Using ORCA, as Figure \ref{fig:architectures} shows, decoder-only models do not outperform their randomly-initialized counterparts for any of the tasks, while encoder-only models do so for all of the tasks, at least with one of the tested models. On the other hand, when using FPT, both encoder-only and decoder-only models do outperform their randomly-initialized versions for most of the tasks (except \textsc{Pythia} for Navier Strokes).
However, both sets of models still show very large error compared to ORCA-based adaptation, and in some cases (for example, \textsc{Pythia} for Diffusion-Reaction), the performance gain is small.
We contend that applying these approaches should only be done when the pre-training in the original modality is necessary; otherwise, there is no gain from pre-training a model at all.

When comparing the performance pre-trained models of different architectures, as Figure \ref{fig:architectures} shows, \textbf{encoder-only models outperform decoder-only models overall} for three of the four selected tasks (Advection, Diffusion-Reaction, and Navier-Stokes), with very different performance depending on the task.
The remaining task, Diffusion-Sorption, shows equally good performance for all models and cross-modal adaptation methods, indicating that the task is simple enough to be solved without pre-training.
Similarly to what \citet{de2024explains} report with the Satellite dataset for satellite image time series analysis, this highlights the importance of selecting tasks that allow us to better evaluate cross-modal adaptation methods.
Broadly, we also observe that ORCA achieves better results than FPT on three of the four tasks, as previously described in \citet{shen2023cross}.

Lastly, it must be noted that for some of these tasks, there is large variance between runs.
For ORCA, all tasks are stable except for Navier-Stokes, where we can see high variance when using encoder-only models that have been randomly initialized.
When using their pre-trained counterparts, this variance reduces dramatically.
On the other hand, when using FPT, all tasks except for Advection seem stable.
For Advection, pre-trained encoder-only models show high variance between runs.
As we discuss later, this fine-tuning instability---which could come from optimizers or simply bad regions in weight space---should be investigated more systematically in future work.

Overall, our results show that \textbf{decoder-only models cannot compete with encoder-only models for PDE tasks using cross-modal adaptation methods out of the box}.
In the following sections, we try several approaches to close the performance gap between architectures.

\section{Scaling Decoder-Only Models does not Improve Performance}\label{sec:scaling}

The previous results motivated us to find potential ways in which decoder-only models can achieve comparable performance to encoder-only models.
Since the compared models in the last section were all of similar size, in this section, we test scaling decoder-only models to see if this improves performance, as seen in other areas \citep{kaplan2020scaling,caillaut-etal-2024-scaling,cai2025exploring}.

\begin{figure*}[ht]
\begin{center}
\includegraphics[width=1.0\textwidth]{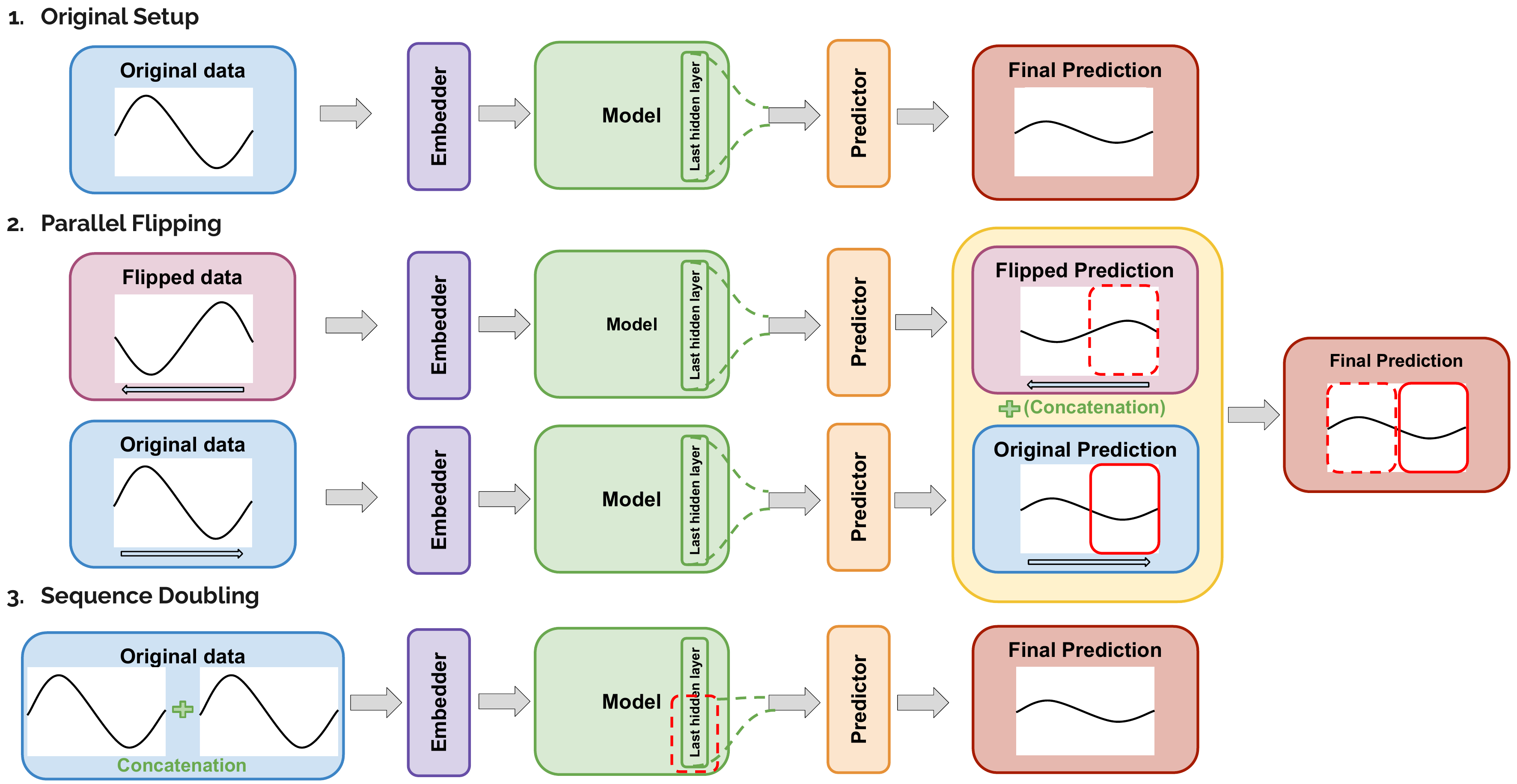}
\end{center}
\caption{Pipeline comparison of the original setup and the two methods we introduce, Parallel Flipping and Sequence Doubling. For \textbf{Parallel Flipping}, the pipeline is run twice, with the original data and with the inverted sequences. For \textbf{Sequence Doubling}, each sequence is concatenated with itself before being introduced to the model, and then we only pass the second part of the last hidden layer to the predictor.}
\label{fig:methods-diagram}
\end{figure*}

However, Figure \ref{fig:scaling} shows that \textbf{scaling barely reduces the performance gap between decoder-only and encoder-only models}, where \textsc{RoBERTa-BASE} represents encoder-only model performance.
Below, we outline the trends we observe.

When using ORCA, there is no performance improvement on the Advection and Diffusion Sorption datasets;
for Advection, there is even some deterioration for both model families.
Diffusion-Reaction shows some improvement with the \textsc{Pythia} models, with some outliers, but no improvement with the \textsc{GPT-2} models.
On the other hand, Navier-Stokes shows some improvement for both model families; the relative percentage improvement when comparing the best model with the smallest model of each family is 5\% for the \textsc{GPT-2} models versus 12\% for the \textsc{Pythia} models.
Still, the trend is not smooth, and the parameter increase to achieve this performance is much bigger than the performance gain;
the best \textsc{GPT-2} model is approximately 12 times bigger than the original size, and the best \textsc{Pythia} model approximately 71 times bigger, without getting much closer to encoder-only model performance.

With FPT-based adaptation, \textsc{GPT-2} models do not show consistent performance improvements, even deteriorating for Advection and Navier-Stokes.
On the other hand, the \textsc{Pythia} family shows some improvements for Advection and Navier-Stokes, but once again, the gains are relatively small compared to the models' size difference.

Since scaling does not close the performance gap between architectures, we hypothesize that the stark differences in performance from plugging decoder-only models into these approaches are due to two reasons: First, they are penalized for being autoregressive, since each point in the sequence is treated as an individual token, and \textsc{GPT-2} and \textsc{Pythia} cannot condition on the sequence bidirectionally, which is necessary for waveforms with symmetry.
Secondly, the predictions are not computed generatively, but instead, the representations of the last hidden layer are simply averaged.
This does not take advantage of the strong generative capabilities that decoder-only models possess. 

In the following sections, we focus on addressing our first hypothesis as a means to improve the performance of decoder-only models using cross-modal adaptation approaches, and leave the exploration of generation for cross-modal adaptation to future work.

\begin{figure*}[ht]
\begin{center}
\includegraphics[width=1.0\textwidth]{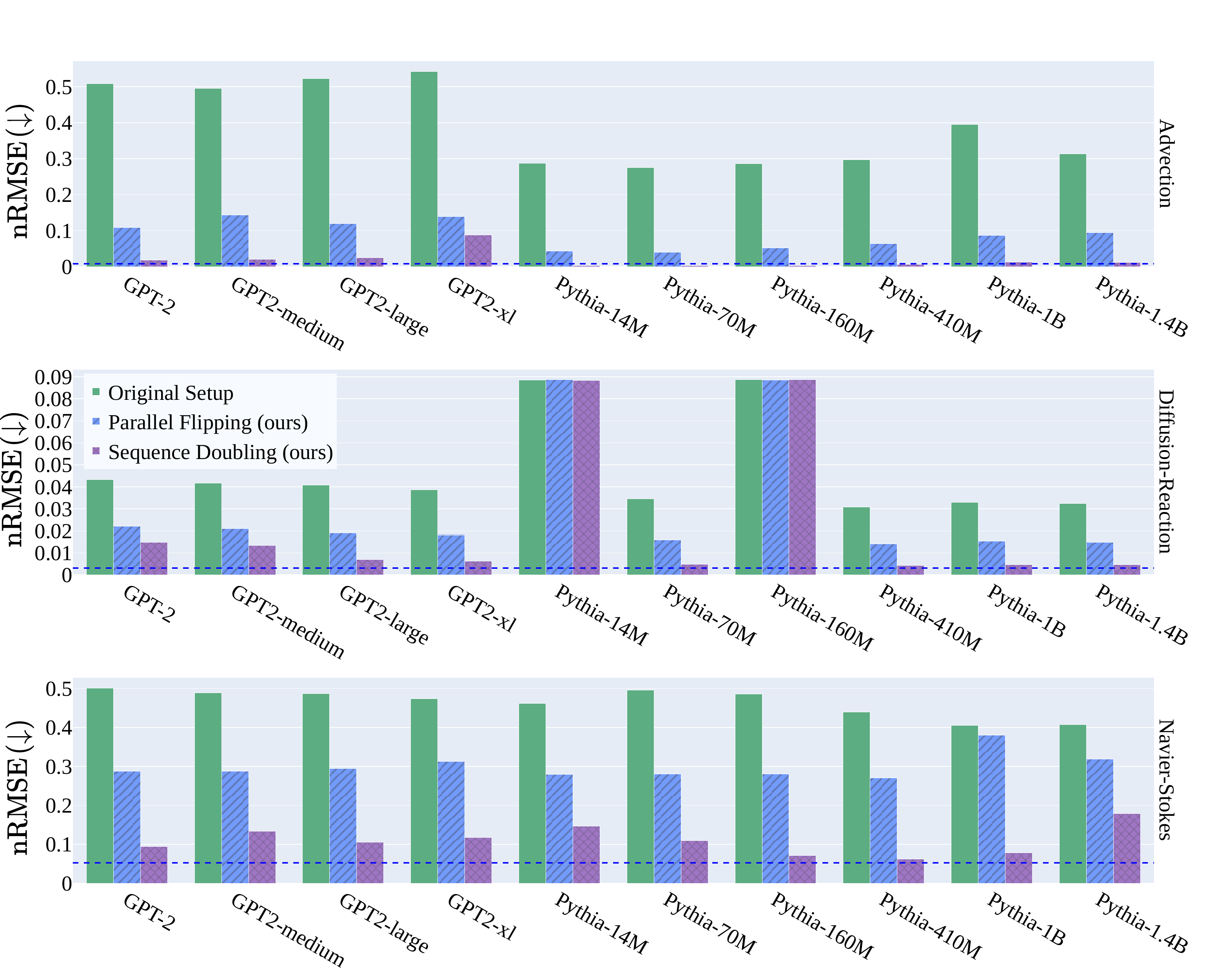}
\end{center}
\caption{Performance comparison of the original setup versus our own two methods, Parallel Flipping and Sequence Doubling, using ORCA \citep{shen2023cross}. We set RoBERTa with the original setup as a baseline for all the configurations. The plots depict the average performance over five random seeds. %
Performance is measured using nRSME, where lower is better. The performance comparison using and FPT \citep{lu2022frozen} can be found in Appendix \ref{sec:appendix-fpt}.}
\label{fig:flipvsdouble}
\end{figure*}

\section{Simulating Bidirectionality With Decoder-Only Models}
\label{sec:methods}

Since scaling decoder-only models does not improve their performance, we introduce two novel methods to counter the lack of bidirectional context in the models, illustrated in Figure \ref{fig:methods-diagram}.

\subsection{Parallel Flipping}

Through error analysis of the decoder-only model outputs, we observed that the beginnings of the output sequences were generally more spiky but they became smoother as the sequence progressed, since the model has more context to condition on.
We show some examples of this in Appendix \ref{sec:error}.

Using this to our advantage, we design a new method to give both halves of the sequence equal opportunity to condition on the other.
As shown in Figure \ref{fig:methods-diagram}, we run the same cross-modal pipeline twice in parallel (both for ORCA and FPT), once with the original data and once with the data sequences inverted.
Then, we combine both predictions by taking the second half of each from the original run and the inverted one and concatenating them.
In this way, both parts of the predicted sequence have access to the previous context and we obtain the smoother part of both runs, even though the point at which they are concatenated can still be spiky.
Compared to the original cross-modal adaptation approach, the second half of the final prediction remains unchanged with Parallel Flipping, but the first half may now improve through conditioning on the flipped version.

\subsection{Sequence Doubling}

To expand the context window the model can use beyond half the sequence (as in Parallel Flipping) to the full sequence, we introduce sequence doubling.
As shown in Figure \ref{fig:methods-diagram}, we concatenate all the sequences introduced to the model with themselves.
Then, for the prediction, we take only the second half of the last hidden layer and introduce it to the predictor.
This half of the hidden layer is conditioned on the first instance of the entire sequence and should therefore be a much richer representation of the data.
Indeed, a similar kind of sequence repetition has also shown promising results in the context of language model embeddings \citep{springer2025repetition}.
Compared to Parallel Flipping, this approach also does not have a hard concatenation point, which should result in smoother and better outputs overall, leading to bigger improvements.
However, Sequence Doubling also doubles the sequence length of inputs, so we reduce the batch size for some configurations.

\section{Simulating Bidirectionality Closes The Performance Gap Between Architectures} %
\label{sec:results}

We compare our two newly-proposed methods with the original setup in Figure \ref{fig:flipvsdouble}, using 
the results of our two new methods in the following section, compared to the original setup.
As Diffusion-Sorption shows equally good performance for all configurations in previous sections, we focus on the other three datasets for these remaining experiments.

Overall, both \textbf{Parallel Flipping and Sequence Doubling outperform the original setup} for all tasks and cross-modal adaptation methods.
As expected, we also see that Sequence Doubling generally outperforms Parallel Flipping, with the exception of FPT-based adaptation of \textsc{GPT-2} models to Advection and Navier-Stokes.

For ORCA, we see larger improvements, with Sequence Doubling outperforming Parallel Flipping for all tasks.
For both Navier-Stokes and Diffusion-Reaction, the performance approaches \textsc{RoBERTa-BASE}'s original performance, and in some cases, our approaches even outperform \textsc{RoBERTa-BASE}, i.e., some \textsc{Pythia} models on Advection (\textsc{Pythia-14M}, \textsc{Pythia-70M}, \textsc{Pythia-160M}, and \textsc{Pythia-410M}).

For FPT (see Appendix \ref{sec:appendix-fpt}), improvements are less consistent than for ORCA, but still show good gains across most configurations.
All models approach \textsc{RoBERTa-BASE}'s performance on Diffusion-Reaction with Sequence Doubling, but the improvement with Parallel Flipping is smaller as before.
For some configurations, decoder-only models outperform encoder-only models again (e.g., \textsc{Pythia} on Advection and Navier-Stokes with Sequence Doubling).
In contrast, \textsc{GPT-2} models show small gains.

Despite the improvements, we still do not consistently see neat scaling behaviour on cross-modal adaptation.
This could be due to a task inherently not benefiting from scaling, particularly for datasets that can already be solved with a lower-capacity model.
On the other hand, the lack of clear scaling could also come from randomness in the particular model checkpoints that we use, which could also cause some of the outlier runs that we see.
Adaptation stability is therefore an important area for future work in cross-modal adaptation.

\section{Discussion and Future Work}
\label{sec:discussion}

With a series of experiments to analyze the effect of model architecture and size on cross-modal adaptation approaches, we show that decoder-only models are consistently worse than encoder-only models and do not, at least with traditional approaches, exploit the potential of their pre-trained knowledge for the new tasks.
We show that this is due to decoder-only models being penalized for their autoregressive attention over the input. To address this penalization, we introduce two different methods, both of which come with certain tradeoffs. %

First, \textbf{Parallel Flipping} requires each instance to be run twice to obtain the final prediction, but by design it can be parallelized, either using double the resources to run in the same time or running it sequentially in double the time. 

On the other hand, \textbf{Sequence Doubling} cannot be parallelized. Also, since the sequence length is doubled, it takes longer to run and increases the required memory. In some cases, particularly when using bigger models, this requires reducing the batch size or upgrading our resources.

Our primary motivation with both methods was to try to mimic the data processing of encoder-only models while using decoder-only models. We did so by introducing a kind of bidirectional context.
Another way of achieving this would be to actually enable bidirectional attention in decoder-only models, as in LLM2Vec \citep{behnamghader2024llmvec}, or by merging encoder-only and decoder-only models as in \citep{charpentier-samuel-2024-bert}. We leave this to future work.

We also notice that cross-modal adaptation methods are very sensitive to the task, with performance varying greatly between the four PDEs we consider.
To better assess the utility of methods in this setting, future work should investigate which properties of PDEs are responsible for this variance.

However, we see the most important direction for future work as being to diagnose the instabilities of cross-modal adaptation, given the high variance of performance with some configurations.
As we point out in Section \ref{sec:architecture}, optimizers might play a role \citep{kunstner2023noise}, as might randomness in the checkpoints we begin with.
One approach would be to try to disentangle when transfer capabilities emerge for these models (and whether that is stable), particularly decoder-only models, and the influence that they have on the variation \citep{van2025polypythias}.

\section{Conclusion}

In this paper, we aim to understand and minimize the impact of model architecture and size on cross-modal adaptation approaches with time-dependent simulation of partial differential equations.
We find that decoder-only models perform much worse than encoder-only models, even when scaled up.
Unidirectional attention plays a key role in this performance gap, preventing models from conditioning on the data overall.
To mitigate the effects of the lack of bidirectionality, we introduce two novel approaches: Parallel Flipping and Sequence Doubling, both of which outperform the original setup, with Sequence Doubling showing much larger gains and closing the gap to encoder-only model performance.
We encourage future research on scientific machine learning to build on our approach to leverage more capable decoder-only models in cross-modal adaptation research.

\section{Limitations}

We only experiment with two popular cross-modal adaptation methods, and leave it to future work to investigate whether the same patterns hold for PARE~\citep{pmlr-v235-cai24c} and UPS~\citep{shen2024ups}.

Our experiments focus on 1-dimensional PDE datasets and tasks widely used in the field of cross-modal adaptation \citep{lu2022frozen,shen2023cross,pmlr-v235-ma24d,shen2024ups,pmlr-v235-cai24c}, even though different PDE tasks (like higher-dimensional PDEs) and PDE-specific evaluations (e.g., with physics-based metrics) could also be used.
We caution against claims about generalization of our results on other PDE tasks and evaluations.

Additionally, given our difficulties replicating the original proxy dataset from ORCA \citep{shen2023cross}, more testing is required to determine the potential influence this could have on all models.

\section{Ethics Statement}

All datasets and models are used in accordance with their licenses and intended use.

\section{Reproducibility statement}

Our code is available here: \url{https://github.com/palomagh/DecodingPDEs}

We use Nvidia A100 GPUs to run all experiments. We perform five runs per configuration, using different random seeds, for a total of $232$ different configurations. The running times spanned from $12$ to $140$ GPU-hours, depending on the dataset and model size.

\bibliography{iclr2026_conference}
\bibliographystyle{iclr2026_conference}

\appendix

\section{PDE Datasets Details and Configurations}\label{sec:appendix-pde}

As we saw in Section \ref{sec:setup}, we tested the models in a collection of PDE datasets from PDEBench \citep{takamoto2022pdebench}. We follow
\citet{shen2023cross} for the download, pre-processing, and loading of the data. %

The specifications of the selected datasets are shown in Table \ref{tab:pde}.

\begin{table*}[h]
    \caption{List of PDE datasets used as target datasets and their corresponding specifications.}
    \label{tab:pde}
    \begin{center}
    \begin{adjustbox}{width=1\textwidth}
    \begin{tabular}{cccccc}
        \toprule
        Dataset & Dimension &  Resolution & Coefficients & Optimizer & \\
        \midrule
        Advection & 1D & 1024 & $\beta = 0.4$ & Adam & \\
        Diffusion-Reaction & 1D & 1024 & $\nu = 0.5, \rho = 1.0$ & SGD & \\
        Diffusion-Sorption & 1D & 1024 &  - & AdamW &  \\
        Compressible Navier-Stokes & 1D & 1024 & $\eta = \zeta = 0.1$, rand periodic & AdamW & \\
        \bottomrule
    \end{tabular}
    \end{adjustbox}
    \end{center}
\end{table*}

\subsection{1D Advection Equation}

The Advection equations are defined as follows:

\begin{equation}
\partial_{t} u(t, x) + \beta \, \partial_{x} u(t, x) = 0
\end{equation}
\begin{equation}
u(0, x) = u_{0}(x)
\end{equation}

where $\beta$ represents a given advection speed and $x \in (0,1)$ and $t \in (0,2]$. 

\subsection{1D Diffusion-Reaction Equation} 

The Diffusion-Reaction equation is expressed as: 

\begin{equation}
    \partial_{t} u(t,x) - \nu \partial_{xx} u(t,x) = \rho u(t,x) (1 - u(t,x))
\end{equation}

where $x \in (0,1)$ and $ t \in (0,1]$.

\subsection{1D Diffusion-Sorption Equation}

The Diffusion-Sorption equation is defined as follows: 
\begin{equation}
    \partial_{t}u(t,x) = \frac{D}{R(u)} \partial_{xx}u(t,x)
\end{equation}

where $x \in (0,1)$, $t \in (0,500]$, and $ D = 5 \times 10^{-4}$. We follow \citet{takamoto2022pdebench} for the details of the function $R(u)$, describing the external force slowing down the diffusion process.

\subsection{1D Compressible Navier-Stokes Equation}

The Compressible Navier-Strokes equations are expressed as:

\begin{equation}
    \partial_{t}p + \nabla \cdot (\rho u) = 0,
\end{equation}
\begin{equation}
    \rho (\partial_{t}u + u \cdot \nabla u) = - \nabla p + \eta \Delta u + \left(\xi + \frac{\eta}{3}\right) \nabla (\nabla \cdot u)
\end{equation}
\begin{equation}
    \partial_{t} \left( \epsilon + \rho \frac{\|u\|_{2}^{2}}{2} \right) + \nabla \cdot \left( \left(p + \epsilon + \rho \frac{u^{2}}{2}\right)u+u \cdot \sigma'\right) = 0
\end{equation}

where $\rho$ represents the density, $u$ represents the velocity, $p$ represents the pressure, and $\epsilon$ the internal energy of the system. Here, $x \in (-1,1)$, and $t \in (0,1)$.

\section{Proxy Datasets}\label{sec:appendix-proxy}

To create proxy datasets for \textsc{GPT-2}, \textsc{GPT-2 Medium} , \textsc{GPT-2 Large} , and \textsc{GPT-2 XL}, we follow the approach detailed in \citet{shen2023cross}. 
Due to discrepancies between the stated dataset and instructions in \citet{shen2023cross}, we use the CoNLL-2000 dataset \citep{sang2000introduction} instead of CoNLL-2003.
We select a random sample of 2000 sequences containing less than 32 tokens.
We unify the length by padding to a sequence length of 32.
Lastly, we calculate the embeddings using the selected models.
\pagebreak

\section{Rescaled Results for Diffusion-Reaction and Diffusion-Sorption}
\label{sec:appendix-zoomin}

We show results from Figure \ref{fig:architectures} and \ref{fig:scaling} for Diffusion-Reaction and Diffusion-Sorption at a smaller scale for better readability.

\begin{figure}[h]
    \centering
    \includegraphics[width=\linewidth]{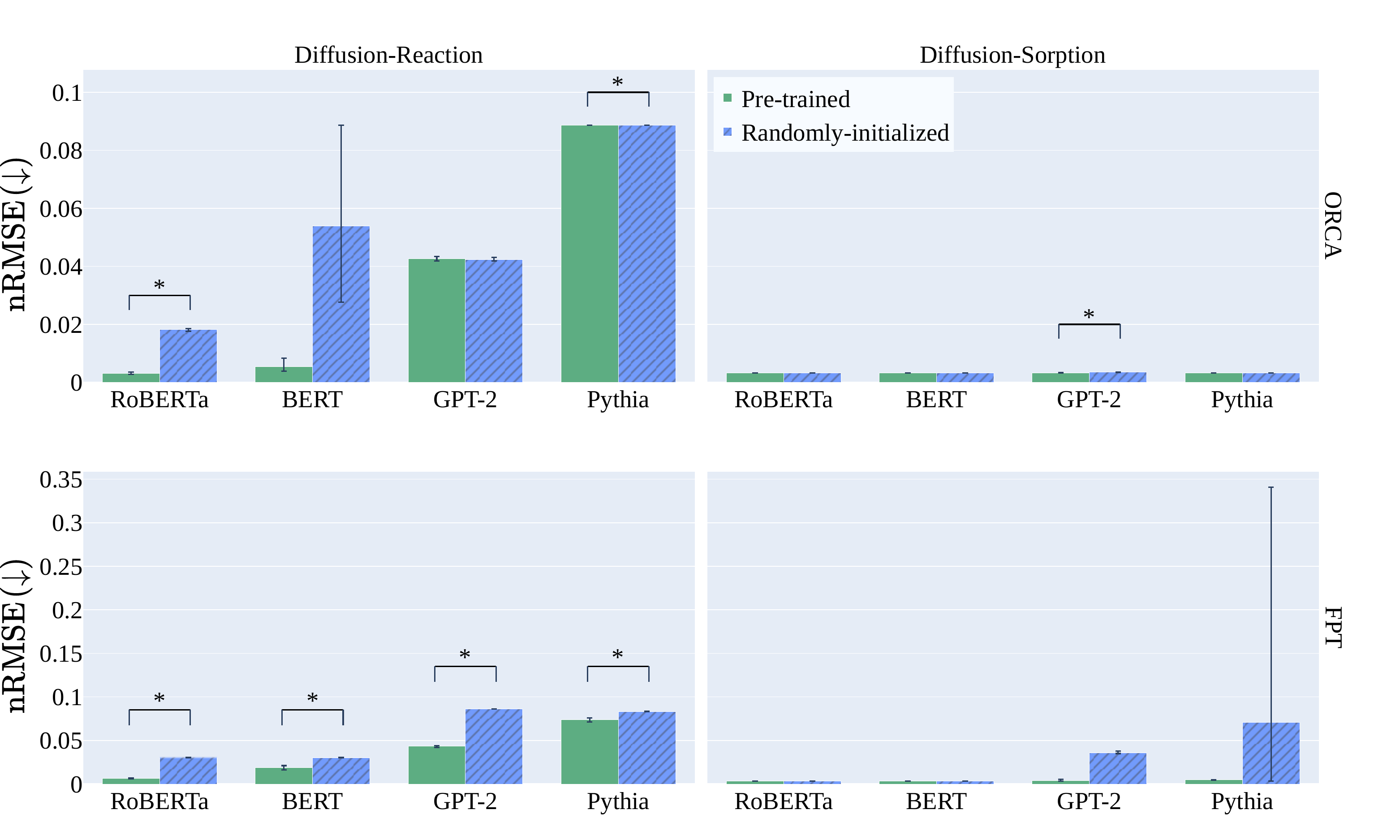}
    \caption{Comparison of model performance with ORCA- (above) and FPT-based (below) cross-modal adaptation, using both pre-trained and randomly-initialized versions of encoder-only models (\textsc{RoBERTa}, \textsc{BERT}) and decoder-only models \textsc{GPT-2} and \textsc{Pythia}). Performance is measured using nRSME, where lower is better; the plots show average performance over five random seeds, and the error bars represent the best and worst runs. We denote with an asterisk ($\ast$) the cases in which there is a statistically significant difference between the pre-trained model and the randomly initialized one; a detailed explanation can be found in Appendix \ref{sec:appendix-statistic}.}
    \label{fig:rd-ds-architecture}
\end{figure}

\begin{figure}[h]
    \centering
    \includegraphics[width=\linewidth]{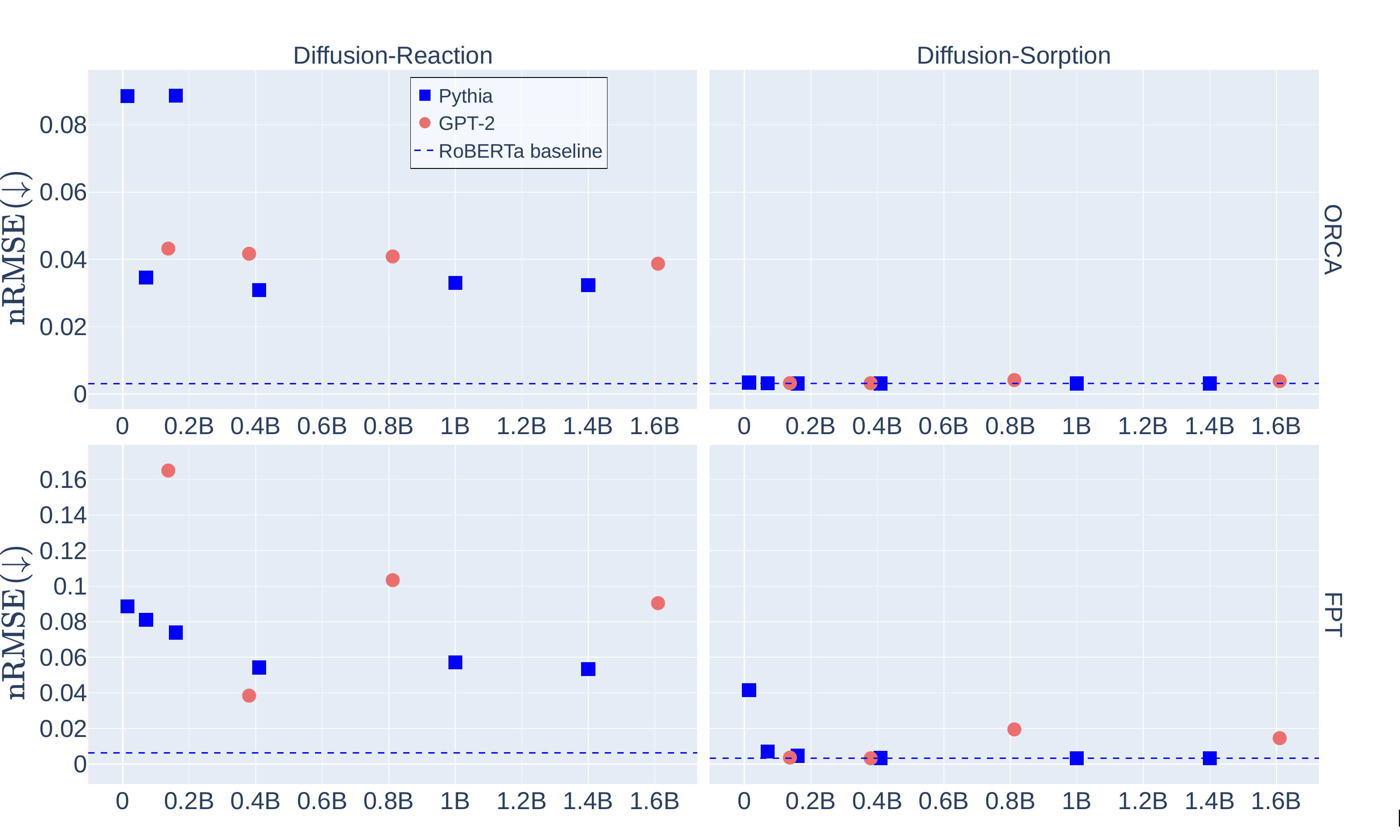}
    \caption{Performance of different sizes of models of the \textsc{GPT-2} family and \textsc{Pythia} family using both ORCA \citep{shen2023cross} and FPT \citep{lu2022frozen}. The plots depict the average performance over five random seeds. Once again, performance is measured using nRSME, where lower is better. If scaling the models was improving the performance, downward trends could've been seen for the different model families.}
    \label{fig:rd-ds-scaling}
\end{figure}

\section{Error Analysis Examples}
\label{sec:error}

\begin{figure}[h]
    \centering
    \includegraphics[width=1\linewidth]{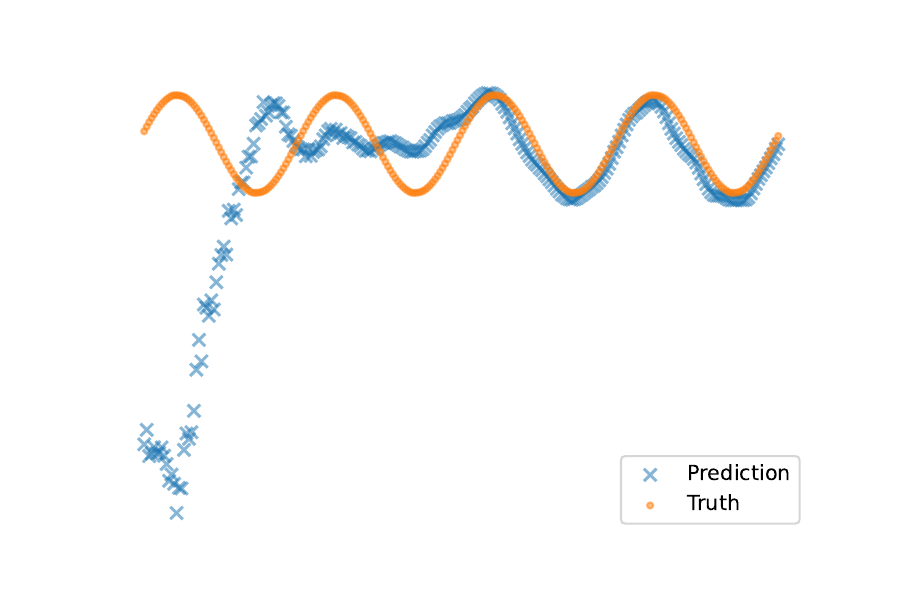}
    \caption{Comparison between \textsc{GPT-2} prediction on an Advection example using ORCA as the cross-modal adaptation method and the ground truth.}
    \label{fig:error1}
\end{figure}

\begin{figure}[h]
    \centering
    \includegraphics[width=1\linewidth]{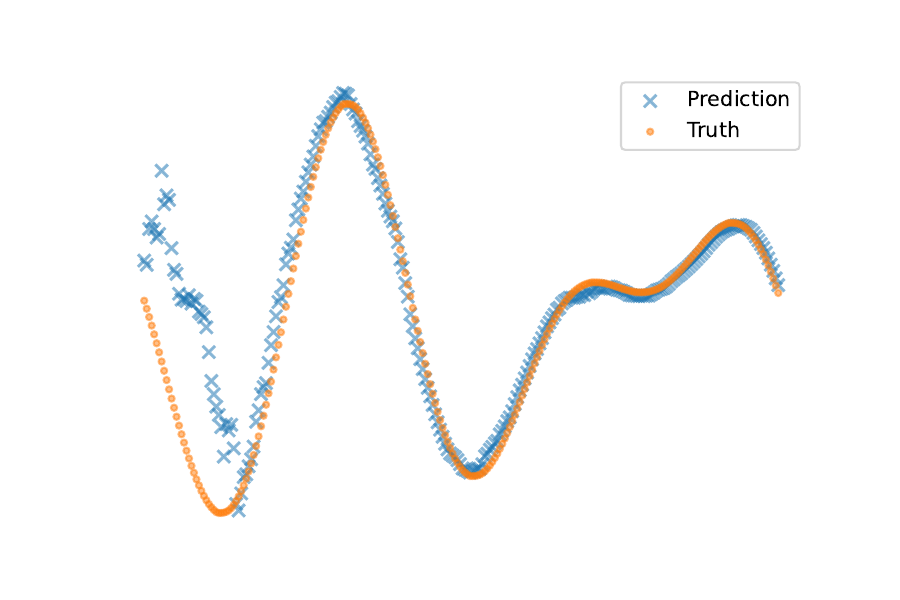}
    \caption{Comparison between \textsc{Pythia} prediction on an Advection example using ORCA as the cross-modal adaptation method and the ground truth.}
    \label{fig:error2}
\end{figure}

We show the comparison of the predicted waves for different examples of Advection and Diffusion-Reaction with different models and the ground truth wave. In Figures \ref{fig:error1} and \ref{fig:error2} we can see that the predictions are more spiky and irregular in the first half of the wave than in the second half. 

\section{Our Methods with Frozen Pre-Trained Transformers}
\label{sec:appendix-fpt}

In this appendix, we include the complementary plot to Figure \ref{fig:flipvsdouble}, using FPT.

\begin{figure*}[ht]
\begin{center}
\includegraphics[width=1.0\textwidth]{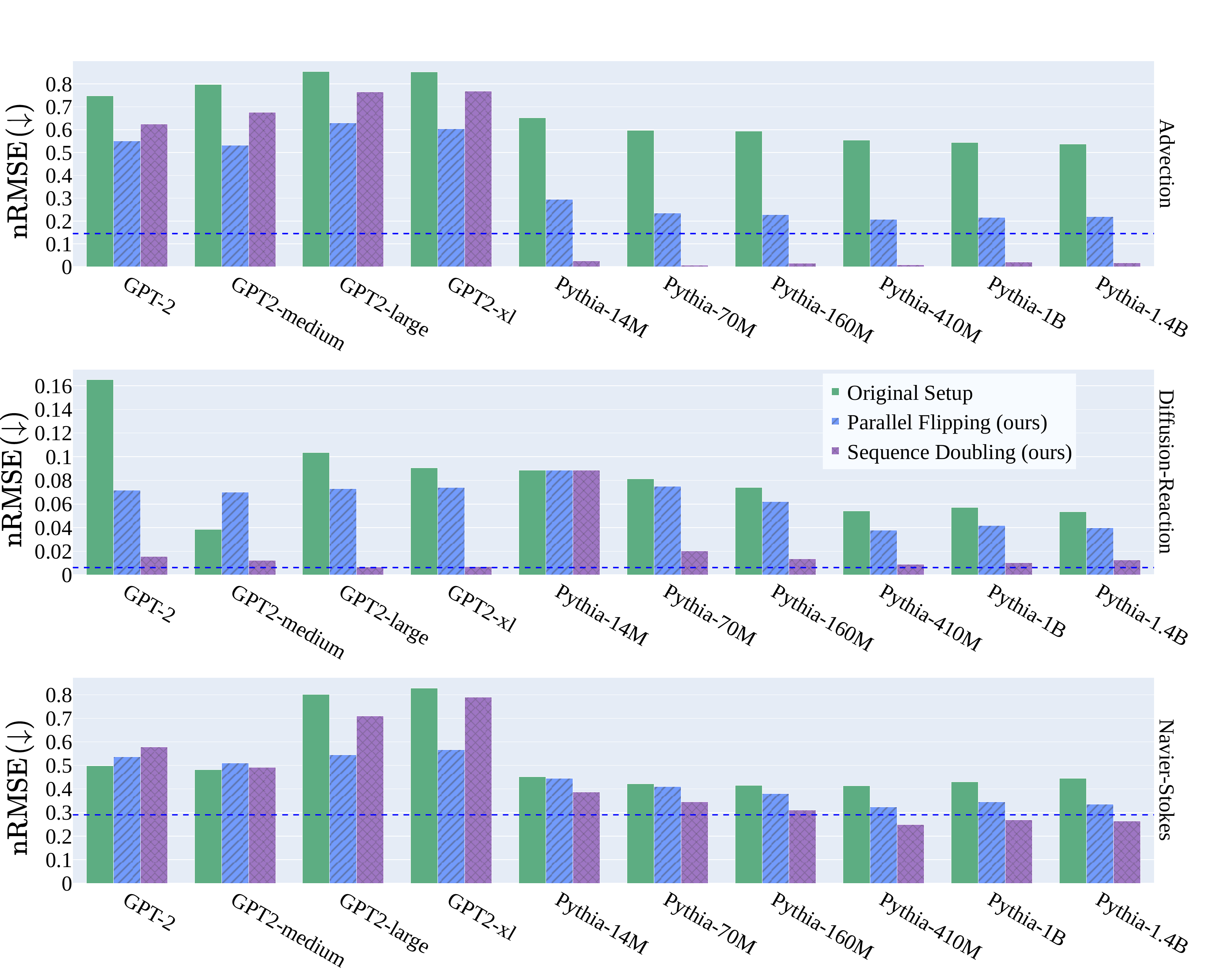}
\end{center}
\caption{Performance comparison of the original setup versus our own two methods, Parallel Flipping and Sequence Doubling, using FPT \citep{lu2022frozen}. We set \textsc{RoBERTa} with the original setup as a baseline for all the configurations. The plots depict the average performance over five random seeds. %
Performance is measured using nRSME, where lower is better.}
\label{fig:flipvsdouble-FPT}
\end{figure*}

\section{Complete Results}
\label{sec:appendix-results}

In this appendix, we include the results of all the runs performed. Tables \ref{tab:random1} and \ref{tab:random2} include the runs comparing the performance of the encoder-only models (\textsc{RoBERTa}, \textsc{BERT})  versus the decoder-only  \textsc{GPT-2} and \textsc{Pythia}), both pre-trained and randomly-initialized. Tables \ref{tab:ADV}, \ref{tab:RD}, and \ref{tab:1DCFD} include the comparison of the different sizes of the decoder-only models (\textsc{GPT-2} and \textsc{Pythia}), with the original setup and both of our newly introduced methods (Parallel Flipping and Sequence Doubling). 

\begin{table}
    \caption{Model performance of all five runs with ORCA- and FPT-based cross-modal adaptation, using both pre-trained and randomly-initialized versions of encoder-only models (\textsc{RoBERTa}, \textsc{BERT}) and decoder-only models \textsc{GPT-2} and \textsc{Pythia}) for Advection and Diffusion-Reaction.}
    \label{tab:random1}
    \begin{center}
    \begin{adjustbox}{width=1\textwidth}
    \begin{tabular}{cccccc}
        \toprule
         \multirow{2}{*}{Methods} & \multirow{2}{*}{Models} & \multicolumn{2}{c}{Advection} & \multicolumn{2}{c}{Diffusion-Reaction}  \\
         & & Pre-Trained & Random & Pre-Trained & Random \\
         \midrule
         \multirow{4}{*}{ORCA} & RoBERTa & \textbf{8,43E-03 $\pm$ 2,86E-04} & 1,28E-02 $\pm$ 2,40E-03 & \textbf{3,11E-03 $\pm$ 2,97E-04} & 1,82E-02 $\pm$ 3,51E-04 \\
         & BERT & 1,80E-02 $\pm$ 5,08E-03 & 1,54E-02 $\pm$ 6,42E-04 & 5,40E-03 $\pm$ 1,74E-03 & 5,38E-02 $\pm$ 3,19E-02 \\
         & GPT-2 & 5,08E-01 $\pm$ 6,94E-03 & 4,89E-01 $\pm$ 5,05E-03 & 4,32E-02 $\pm$ 5,11E-04 & 4,24E-02 $\pm$ 6,40E-04  \\
         & Pythia & 2,86E-01 $\pm$ 3,90E-03 & 3,11E-01 $\pm$ 5,96E-03 & 8,86E-02 $\pm$ 2,15E-06 & 8,86E-02 $\pm$ 1,03E-05 \\
         \midrule
         \multirow{4}{*}{FPT}& RoBERTa & 1,45E-01 $\pm$ 1,59E-01 & 8,19E-01 $\pm$ 5,75E-02 & 6,27E-03 $\pm$ 4,48E-04 & 3,03E-02 $\pm$ 1,24E-04  \\
         & BERT & 1,89E-01 $\pm$ 9,98E-02 & 6,09E-01 $\pm$ 5,31E-02 & 1,88E-02 $\pm$ 1,74E-03 & 3,03E-02 $\pm$ 1,92E-04  \\
         & GPT-2 & 7,48E-01 $\pm$ 1,00E-02 & 1,00E+00 $\pm$ 2,73E-02 & 1,65E-01 $\pm$ 7,33E-02 & 8,61E-02 $\pm$ 1,06E-04  \\
         & Pythia & 5,93E-01 $\pm$ 6,60E-03 & 6,56E-01 $\pm$ 1,31E-03 & 7,39E-02 $\pm$ 1,98E-03 & 8,44E-02 $\pm$ 1,02E-04\\
         \bottomrule
    \end{tabular}
    \end{adjustbox}
    \end{center}
\end{table}

\begin{table}
    \caption{Model performance of all five runs with ORCA- and FPT-based cross-modal adaptation, using both pre-trained and randomly-initialized versions of encoder-only models (\textsc{RoBERTa}, \textsc{BERT}) and decoder-only models \textsc{GPT-2} and \textsc{Pythia}) for Diffusion-Sorption and Navier-Stokes.}
    \label{tab:random2}
    \begin{center}
    \begin{adjustbox}{width=1\textwidth}
    \begin{tabular}{cccccc}
        \toprule
         \multirow{2}{*}{Methods} & \multirow{2}{*}{Models} & \multicolumn{2}{c}{Diffusion-Sorption} & \multicolumn{2}{c}{Navier-Stokes}  \\
         & & Pre-Trained & Random & Pre-Trained & Random \\
         \midrule
         \multirow{4}{*}{ORCA} & RoBERTa & \textbf{3,18E-03 $\pm$ 2,82E-06} & \textbf{3,18E-03 $\pm$ 7,57E-07} & \textbf{5,22E-02 $\pm$ 7,92E-03} & 7,74E-01 $\pm$ 3,84E-01 \\
         & BERT & \textbf{3,18E-03 $\pm$ 1,28E-06} & 3,19E-03 $\pm$ 7,75E-06 & 9,89E-01 $\pm$ 5,04E-03 & 2,76E-01 $\pm$ 4,01E-01 \\
         & GPT-2 &  3,29E-03 $\pm$ 4,41E-05 & 3,52E-03 $\pm$ 5,18E-05 & 5,01E-01 $\pm$ 1,64E-03 & 4,77E-01 $\pm$ 4,30E-03 \\
         & Pythia & \textbf{3,18E-03 $\pm$ 2,00E-06} & \textbf{3,18E-03 $\pm$ 1,33E-06} & 4,86E-01 $\pm$ 6,54E-03 & 5,03E-01 $\pm$ 4,80E-03 \\
         \midrule
         \multirow{4}{*}{FPT}& RoBERTa & 3,29E-03 $\pm$ 4,26E-05 & 3,20E-03 $\pm$ 8,11E-06 & 2,90E-01 $\pm$ 9,37E-03 & 3,28E-01 $\pm$ 2,40E-03 \\
         & BERT & 3,20E-03 $\pm$ 9,70E-06 & 3,20E-03 $\pm$ 5,54E-06 & 2,82E-01 $\pm$ 4,17E-03 & 3,29E-01 $\pm$ 1,40E-03 \\
         & GPT-2 & 3,62E-03 $\pm$ 8,84E-04 & 3,59E-02 $\pm$ 1,22E-03 & 7,03E-01 $\pm$ 1,79E-02 & 6,49E-01 $\pm$ 1,40E-02 \\
         & Pythia & 4,66E-03 $\pm$ 2,94E-04 & 7,09E-02 $\pm$ 1,51E-01 & 4,15E-01 $\pm$ 7,96E-04 & 4,20E-01 $\pm$ 4,84E-04 \\
         \bottomrule
    \end{tabular}
    \end{adjustbox}
    \end{center}
\end{table}

\begin{table}
    \caption{Model performance of all five runs with ORCA- and FPT-based cross-modal adaptation, with different sizes of decoder-only models \textsc{GPT-2} and \textsc{Pythia}) for \textbf{Advection}, using both the Original setup and our two new approaches.}
    \label{tab:ADV}
    \begin{center}
    \begin{adjustbox}{width=1\textwidth}
    \begin{tabular}{cccccccc}
        \toprule
        \multirow{2}{*}{Model Family} & \multirow{2}{*}{Size} & \multicolumn{3}{c}{ORCA} & \multicolumn{3}{c}{FPT} \\
        & & Original Setup & Parallel Flipping (ours) & Sequence Doubling (ours)  & Original Setup & Parallel Flipping (ours) & Sequence Doubling (ours)  \\
        \midrule
         \multirow{4}{*}{GPT-2} & 137M & 5,08E-01 $\pm$ 6,94E-03 & 1,09E-01 $\pm$ 4,99E-03 & \textbf{1,81E-02 $\pm$ 1,98E-03} & 7,48E-01 $\pm$ 1,00E-02  &  \textbf{5,51E-01 $\pm$ 1,88E-02} & 6,25E-01 $\pm$ 3,90E-02\\
         & 380M & 4,95E-01 $\pm$ 4,07E-03 & 1,44E-01 $\pm$ 8,39E-02 & \textbf{2,04E-02 $\pm$ 3,71E-03} & 7,97E-01 $\pm$ 1,34E-02 & \textbf{5,32E-01 $\pm$ 4,28E-02}  & 6,77E-01 $\pm$ 1,92E-02 \\
         & 812M & 5,23E-01 $\pm$ 1,59E-02 & 1,19E-01 $\pm$ 9,63E-03 & \textbf{2,49E-02 $\pm$ 8,03E-03} & 8,54E-01 $\pm$ 9,31E-03 & \textbf{6,29E-01 $\pm$ 2,28E-02}  & 7,65E-01 $\pm$ 1,71E-02  \\
        & 1.61B & 5,42E-01 $\pm$ 1,01E-02 & 1,39E-01 $\pm$ 9,11E-03 & \textbf{8,83E-02 $\pm$ 6,41E-03} & 8,53E-01 $\pm$ 6,69E-03 & \textbf{5,98E-01 $\pm$ 3,87E-02} & 7,69E-01 $\pm$ 1,96E-02 \\
        \midrule
         \multirow{6}{*}{Pythia} & 14M & 2,87E-01 $\pm$ 3,44E-03 & 4,38E-02 $\pm$ 1,55E-03 & \textbf{2,83E-03 $\pm$ 9,42E-05} & 6,52E-01 $\pm$ 3,72E-03 &  2,95E-01 $\pm$ 6,70E-03 & \textbf{2,54E-02 $\pm$ 1,92E-03} \\
         & 70M & 2,75E-01 $\pm$ 1,68E-02 & 3,98E-02 $\pm$ 1,39E-03 & \textbf{3,02E-03 $\pm$ 1,39E-04} & 5,97E-01 $\pm$ 3,78E-03 & 2,35E-01 $\pm$ 7,53E-03 & \textbf{7,25E-03 $\pm$ 9,94E-04} \\
         & 160M & 2,86E-01 $\pm$ 3,90E-03 & 5,20E-02 $\pm$ 9,22E-04 & \textbf{3,72E-03 $\pm$ 3,51E-04} & 5,93E-01 $\pm$ 6,60E-03 & 2,29E-01 $\pm$ 1,44E-04 & \textbf{1,58E-02 $\pm$ 5,84E-03} \\
         & 410M & 2,97E-01 $\pm$ 3,21E-03 & 6,43E-02 $\pm$ 4,46E-03 & \textbf{6,47E-03 $\pm$ 2,81E-04} & 5,55E-01 $\pm$ 3,37E-03 & 2,07E-01 $\pm$ 3,19E-03 & \textbf{8,88E-03 $\pm$ 2,74E-04}  \\
         & 1B & 3,95E-01 $\pm$ 4,64E-02 & 8,72E-02 $\pm$ 7,45E-03 & \textbf{1,27E-02 $\pm$ 2,95E-04} & 5,43E-01 $\pm$ 6,10E-03 & 2,16E-01 $\pm$ 6,19E-03 & \textbf{2,04E-02 $\pm$ 9,10E-03}\\
         & 1.4B & 3,14E-01 $\pm$ 4,22E-03 & 9,50E-02 $\pm$ 2,10E-02 & \textbf{1,23E-02 $\pm$ 2,08E-04} & 5,37E-01 $\pm$ 8,54E-03 & 2,19E-01 $\pm$ 2,83E-03 & \textbf{1,69E-02 $\pm$ 5,83E-03}  \\
         \bottomrule
    \end{tabular}
    \end{adjustbox}
    \end{center}
\end{table}

\begin{table}
    \caption{Model performance of all five runs with ORCA- and FPT-based cross-modal adaptation, with different sizes of decoder-only models \textsc{GPT-2} and \textsc{Pythia}) for \textbf{Diffusion-Reaction}, using both the Original setup and our two new approaches.}
    \label{tab:RD}
    \begin{center}
    \begin{adjustbox}{width=1\textwidth}
    \begin{tabular}{cccccccc}
        \toprule
        \multirow{2}{*}{Model Family} & \multirow{2}{*}{Size} & \multicolumn{3}{c}{ORCA} & \multicolumn{3}{c}{FPT} \\
        & & Original Setup & Parallel Flipping (ours) & Sequence Doubling (ours) & Original Setup & Parallel Flipping (ours) & Sequence Doubling (ours)  \\
        \midrule
         \multirow{4}{*}{GPT-2} & 137M & 4,32E-02 $\pm$ 5,11E-04 & 2,21E-02 $\pm$ 7,95E-04  & \textbf{1,48E-02 $\pm$ 2,07E-03} & 1,65E-01 $\pm$ 7,33E-02 & 7,18E-02 $\pm$ 6,40E-04  & \textbf{1,57E-02 $\pm$ 1,32E-03}  \\
         & 380M & 4,17E-02 $\pm$ 1,20E-03 & 2,10E-02 $\pm$ 5,83E-04 & \textbf{1,34E-02 $\pm$ 6,08E-04} & 3,85E-02 $\pm$ 8,26E-04 & 6,99E-02 $\pm$ 1,40E-03 & \textbf{1,25E-02 $\pm$ 1,06E-03}\\
         & 812M & 4,30E-02 $\pm$ 1,90E-03 & 1,90E-02 $\pm$ 9,89E-04 & \textbf{7,08E-03 $\pm$ 3,19E-04} & 1,03E-01 $\pm$ 3,91E-02 & 7,29E-02 $\pm$ 4,77E-04 & \textbf{6,85E-03 $\pm$ 4,91E-04 }\\
        & 1.61B & 3,89E-02 $\pm$ 5,93E-04 & 1,81E-02 $\pm$ 3,32E-04 & \textbf{6,25E-03 $\pm$ 3,33E-04} & 9,05E-02 $\pm$ 1,80E-02 & 7,41E-02 $\pm$ 5,38E-04 & \textbf{6,97E-03 $\pm$ 4,26E-04}  \\
         \midrule
         \multirow{6}{*}{Pythia} & 14M & 8,84E-02 $\pm$ 3,07E-04 & 8,86E-02 $\pm$ 5,00E-04 & \textbf{8,83E-02 $\pm$ 3,97E-04} & 8,87E-02 $\pm$ 6,75E-05 & \textbf{8,86E-02 $\pm$ 9,50E-05} & 8,87E-02 $\pm$ 1,24E-04 \\
         & 70M & 3,46E-02 $\pm$ 2,26E-03 & 1,58E-02 $\pm$ 1,60E-03 & \textbf{4,82E-03 $\pm$ 2,75E-04} & 8,11E-02 $\pm$ 6,21E-04 & 7,51E-02 $\pm$ 1,11E-03 & \textbf{2,02E-02 $\pm$ 7,62E-04}\\
         & 160M & \textbf{8,86E-02 $\pm$ 2,15E-06} & 8,85E-02 $\pm$ 1,70E-05 & \textbf{8,86E-02 $\pm$ 2,18E-06} & 7,39E-02 $\pm$ 1,98E-03 & 6,20E-02 $\pm$ 1,75E-03 & \textbf{1,37E-02 $\pm$ 1,12E-03} \\
         & 410M & 3,09E-02 $\pm$ 9,22E-04 & 1,42E-02 $\pm$ 4,95E-04 & \textbf{4,33E-03 $\pm$ 5,96E-04} & 5,43E-02 $\pm$ 1,53E-03 & 3,84E-02 $\pm$ 1,65E-03 & \textbf{9,06E-03 $\pm$ 3,37E-04}\\
         & 1B & 3,30E-02 $\pm$ 1,16E-03 & 1,53E-02 $\pm$ 3,81E-04 & \textbf{4,69E-03 $\pm$ 3,15E-04} & 5,71E-02 $\pm$ 6,68E-03 & 4,18E-02 $\pm$ 7,66E-03 & \textbf{1,04E-02 $\pm$ 1,45E-03} \\
         & 1.4B & 3,24E-02 $\pm$ 8,12E-04 & 1,50E-02 $\pm$ 5,06E-04 & \textbf{4,71E-03 $\pm$ 2,87E-04} & 5,33E-02 $\pm$ 7,01E-03 & 3,99E-02 $\pm$ 9,06E-03 & \textbf{1,25E-02 $\pm$ 2,88E-03} \\
         \bottomrule
    \end{tabular}
    \end{adjustbox}
    \end{center}
\end{table}

\begin{table}
    \caption{Model performance of all five runs with ORCA- and FPT-based cross-modal adaptation, with different sizes of decoder-only models \textsc{GPT-2} and \textsc{Pythia}) for \textbf{Navier-Stokes}, using both the Original setup and our two new approaches.}
    \label{tab:1DCFD}
    \begin{center}
    \begin{adjustbox}{width=1\textwidth}
    \begin{tabular}{cccccccc}
        \toprule
        \multirow{2}{*}{Model Family} & \multirow{2}{*}{Size} & \multicolumn{3}{c}{ORCA} & \multicolumn{3}{c}{FPT} \\
        & & Original Setup & Parallel Flipping (ours) & Sequence Doubling (ours)  & Original Setup & Parallel Flipping (ours) & Sequence Doubling (ours) \\
        \midrule
         \multirow{4}{*}{GPT-2} & 137M & 5,01E-01 $\pm$ 1,64E-03& 2,88E-01 $\pm$ 2,50E-03 & \textbf{9,44E-02 $\pm$ 5,63E-03}& 7,03E-01 $\pm$ 1,79E-02 & \textbf{5,37E-01 $\pm$ 5,59E-03}  & 5,79E-01 $\pm$ 3,59E-02 \\
         & 380M & 4,89E-01 $\pm$ 1,30E-03 & 2,87E-01 $\pm$ 1,53E-03 & \textbf{1,34E-01 $\pm$ 4,92E-03} & \textbf{4,82E-01 $\pm$ 2,35E-03} & 5,10E-01 $\pm$ 8,90E-03 & 4,92E-01 $\pm$ 1,83E-02  \\
         & 812M & 4,87E-01 $\pm$ 2,60E-03 & 2,94E-01 $\pm$ 1,03E-03 & \textbf{1,05E-01 $\pm$ 5,95E-03} & 8,02E-01 $\pm$ 8,25E-03 & \textbf{5,45E-01 $\pm$ 1,11E-02} & 7,10E-01 $\pm$ 4,69E-02\\
        & 1.61B & 4,74E-01 $\pm$ 3,30E-03 & 3,13E-01 $\pm$ 1,32E-02 & \textbf{1,17E-01 $\pm$ 1,25E-02} & 8,29E-01 $\pm$ 1,21E-02 & \textbf{5,67E-01 $\pm$ 1,28E-02} & 7,90E-01 $\pm$ 1,73E-02  \\
         \hline
         \multirow{6}{*}{Pythia} & 14M & 4,62E-01 $\pm$ 6,25E-03 & 2,80E-01 $\pm$ 1,91E-03 & \textbf{1,47E-01 $\pm$ 9,26E-03} & 4,52E-01 $\pm$ 4,80E-04 & 4,45E-01 $\pm$ 7,73E-04 & \textbf{3,87E-01 $\pm$ 1,69E-03} \\
         & 70M & 4,96E-01 $\pm$ 2,04E-03 & 2,81E-01 $\pm$ 1,89E-03 & \textbf{1,09E-01 $\pm$ 4,01E-03} & 4,23E-01 $\pm$ 8,73E-04 & 4,10E-01 $\pm$ 8,61E-04 & \textbf{3,45E-01 $\pm$ 2,29E-03} \\
         & 160M & 4,86E-01 $\pm$ 6,54E-03 & 2,81E-01 $\pm$ 1,34E-03 & \textbf{7,12E-02 $\pm$ 4,92E-03} & 4,15E-01 $\pm$ 7,96E-04 & 3,80E-01 $\pm$ 1,48E-02 & \textbf{3,11E-01 $\pm$  1,26E-02} \\
         & 410M & 4,39E-01 $\pm$ 4,66E-03 & 2,70E-01 $\pm$ 2,24E-03 & \textbf{6,25E-02 $\pm$ 8,78E-03} & 4,14E-01 $\pm$ 9,43E-04 & 3,24E-01 $\pm$ 4,41E-03 & \textbf{2,49E-01 $\pm$ 4,71E-03}\\
         & 1B & 4,05E-01 $\pm$ 7,73E-04 & 3,80E-01 $\pm$ 6,76E-02 & \textbf{7,82E-02 $\pm$ 5,19E-03} & 4,30E-01 $\pm$ 7,57E-03 & 3,45E-01 $\pm$ 8,66E-03 & \textbf{2,69E-01 $\pm$ 6,57E-03}\\
         & 1.4B & 4,07E-01 $\pm$ 1,11E-03 & 3,53E-01 $\pm$ 7,83E-02 & \textbf{1,79E-01 $\pm$ 5,79E-02} & 4,45E-01 $\pm$ 3,36E-03 & 3,34E-01 $\pm$ 1,11E-02 & \textbf{2,64E-01 $\pm$ 8,38E-03}\\
         \bottomrule
    \end{tabular}
    \end{adjustbox}
    \end{center}
\end{table}

\pagebreak

\section{Statistical Significance}\label{sec:appendix-statistic}

To be able to select the proper test for statistical significance, we first performed the Shapiro-Wilk test to assess if the obtained results from the five runs for each configuration were normally distributed. Once we determined this, we utilized two different statistical significance testing methods, the Wilcoxon test (for the cases in which one of the two distributions was not normally distributed) and the t-test (when both distributions were normally distributed), both of them two-sided. For all statistical test ran, we set the $p$-value to 0.05.

\end{document}